\documentclass[letterpaper, 10 pt, conference]{ieeeconf}  

\IEEEoverridecommandlockouts                              

\usepackage{xcolor}
\usepackage{amsmath}
\usepackage{graphicx}
\usepackage{nomencl}
\makenomenclature

\usepackage{listings}
\usepackage{xcolor}

\colorlet{punct}{red!60!black}
\definecolor{background}{HTML}{EEEEEE}
\definecolor{delim}{RGB}{20,105,176}
\colorlet{numb}{magenta!60!black}

\lstdefinelanguage{json}{
    basicstyle=\tiny\ttfamily,
    numbers=left,
    numberstyle=\scriptsize,
    stepnumber=1,
    numbersep=8pt,
    showstringspaces=false,
    breaklines=true,
    frame=lines,
    backgroundcolor=\color{background},
    xleftmargin=2.5em,
    literate=
      {:}{{{\color{punct}{:}}}}{1}
      {,}{{{\color{punct}{,}}}}{1}
      {\{}{{{\color{delim}{\{}}}}{1}
      {\}}{{{\color{delim}{\}}}}}{1}
      {[}{{{\color{delim}{[}}}}{1}
      {]}{{{\color{delim}{]}}}}{1},
}

\title{\LARGE \bf
A Roadmap Towards \\Automated and Regulated Robotic Systems
}

\author{Yihao Liu $^{1}$ and Mehran Armand $^{1,2}$
\thanks{This work was supported by Johns Hopkins internal fundings, National Institute of Arthritis and Musculoskeletal and Skin Diseases (R01AR080315), and National Institute of Biomedical Imaging and Bioengineering (R01EB023939).}
\thanks{$^{1}$
        Department of Computer Science,
        Johns Hopkins University 
        }
\thanks{$^{2}$
        Department of Orthopaedic Surgery, Department of Mechanical Engineering, 
        Johns Hopkins University 
        }
\thanks{Send correspondence to {\tt\small yliu333@jhu.edu}}
\thanks{Some of the figures in this work contain logos and graphics created with BioRender.com. See Acknowledgement section for a full list of these logos and graphics.}
}

\begin{document}

\maketitle
\thispagestyle{empty}
\pagestyle{empty}

\begin{abstract}
The rapid development of generative technology opens up possibility for higher level of automation, and artificial intelligence (AI) embodiment in robotic systems is imminent. However, due to the blackbox nature of the generative technology, the generation of the knowledge and workflow scheme is uncontrolled, especially in a dynamic environment and a complex scene. This poses challenges to regulations in safety-demanding applications such as medical scenes. We argue that the unregulated generative processes from AI is fitted for low level end tasks, but intervention either in the form of manual or automated regulation should happen post-workflow-generation and pre-robotic-execution. To address this, we propose a roadmap that can lead to fully automated and regulated robotic systems. In this paradigm, the high level policies are generated as structured graph data, enabling regulatory oversight and reusability, while the code base for lower level tasks is generated by generative models. Our approach aims the practical transitioning from expert knowledge to regulated action, akin to the iterative processes of study, practice, scrutiny, and execution in human tasks. We identify the generative and deterministic processes in a design cycle, where generative processes serve as a text-based world simulator and the deterministic processes generate the executable system. We propose State Machine Seralization Language (SMSL) to be the conversion point between text simulator and executable workflow control. From there, we analyze the modules involved based on the current literature, and discuss human in the loop. As a roadmap, this work identifies the current possible implementation and future work. This work does not provide an implemented system but envisions to inspire the researchers working on the direction in the roadmap. We implement the SMSL and D-SFO paradigm that serve as the starting point of the roadmap. The code repository of SMSL is available at https://smsl.dev .
\end{abstract}


\nomenclature{AI}{Artificial Intelligence}
\nomenclature{D-SFO*}{Dispatcher-State/Flag/Operation Paradigm}
\nomenclature{ET*}{Expert Text}
\nomenclature{XML}{Extensible Markup Language}
\nomenclature{FSM}{Finite State Machine}
\nomenclature{HIM*}{Hardware Identification Model}
\nomenclature{hFSM}{Hierarchical Finite State Machine}
\nomenclature{IGT}{Image-Guided Therapy}
\nomenclature{LLM}{Large Language Model}
\nomenclature{OL*}{Operation Library}
\nomenclature{OR}{Operating Room}
\nomenclature{PM*}{Proceduralization Model}
\nomenclature{PSM*}{Procedure Serialization Model}
\nomenclature{RMM*}{Robot Manipulation Model}
\nomenclature{SMSL*}{State Machine Serialization Language}
\nomenclature{FSM-G*}{Serialized Finite State Machine Graph}
\nomenclature{YAML}{Yet Another Markup Language}
\nomenclature{WSN}{Wireless Sensor Network}
\nomenclature{dVRK}{da Vinci Research Kit}
\nomenclature{RLHF}{Reinforcement Learning from Human Feedback}
\nomenclature{USMLE}{United States Medical Licensing Examinations}
\nomenclature{AHA}{American Heart Association}
\nomenclature{BLS}{Basic Life Support}
\nomenclature{ACLS}{Advance Cardiovascular Life Support}
\nomenclature{FDA}{US Food and Drug Administration}
\nomenclature{SAM}{Segment Anything Model}
\nomenclature{SB*}{State Branch}
\nomenclature{NLP}{Natural Language Processing}
\nomenclature{SCF*}{State Check Function} 
\nomenclature{IoT}{Internet of Things}

\printnomenclature
* Some terms are existing nomenclature that are already commonly used. The new nomenclature proposed in our work are marked with an asterisk.

\begin{figure}[ht]
    \centering
    \includegraphics[width=0.3\textwidth]{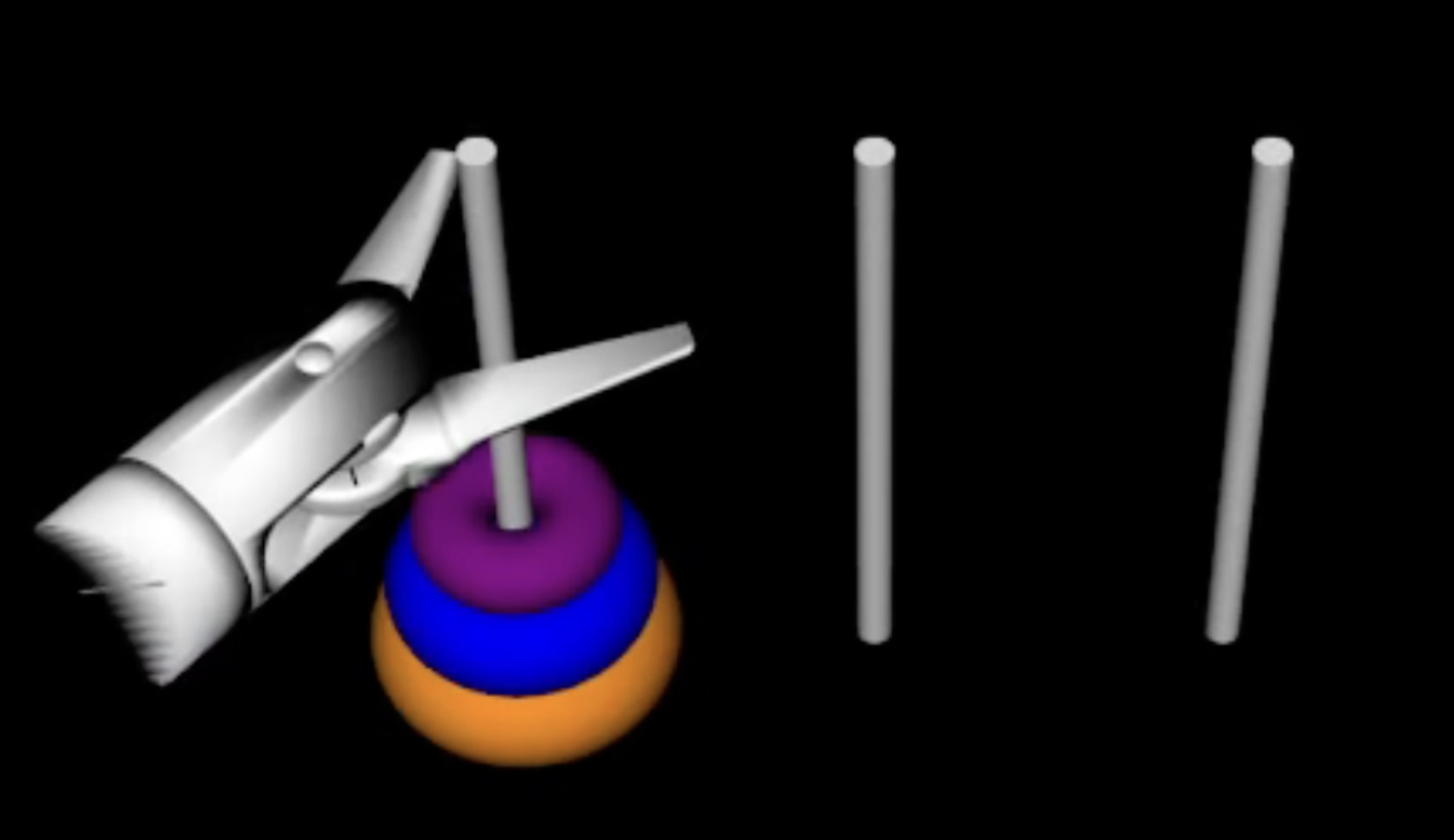}
    \caption{Automatically solving Hanoi Tower using da Vinci Master Tool Manipulator (MTM) in dVRK, with the state transitions controlled by SMSL.}
    \label{fig:davincihanoi}
\end{figure}

\begin{figure*}[ht]
    \centering
    \includegraphics[width=\textwidth]{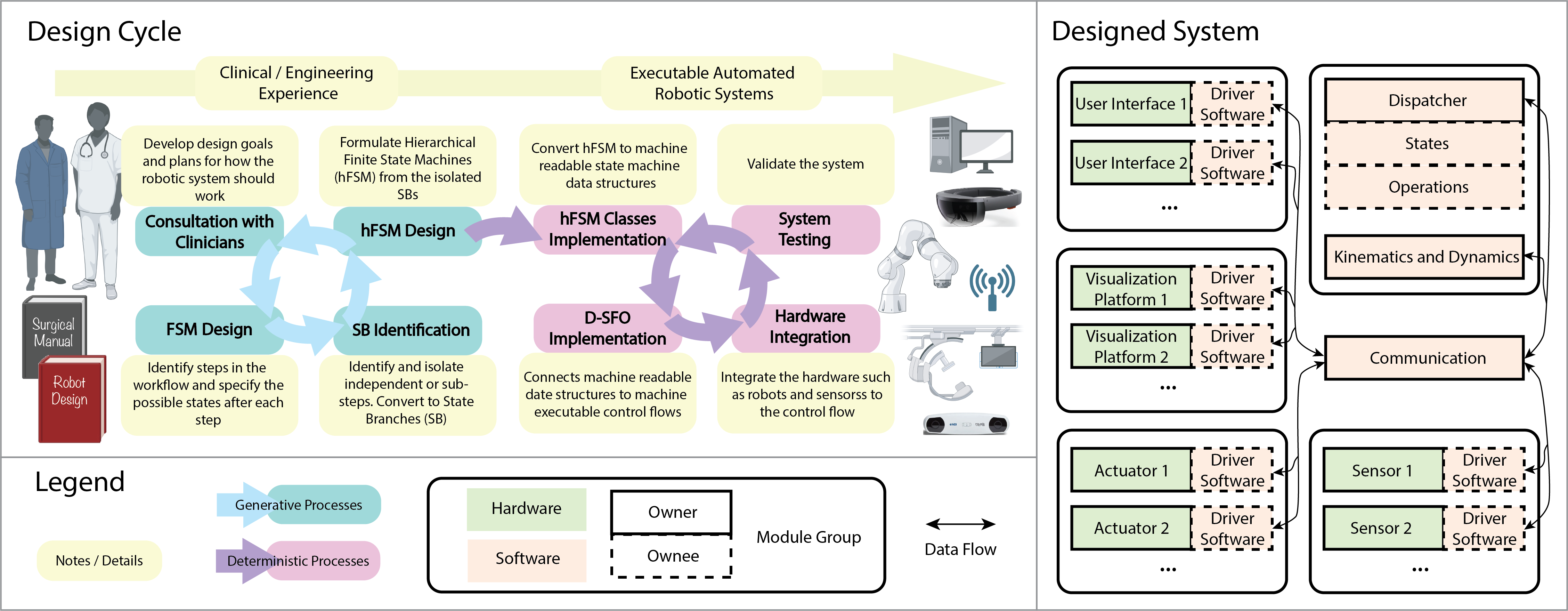}
    \caption{The design cycle and the architecture of a process-controlled medical robotic system \cite{liu2023toward}. The panel \textit{Design Cycle} identifies the generative processes and the deterministic processes in the design cycle. \textit{Consultation with clinicians}, \textit{FSM design}, \textit{SB identification}, and \textit{hFSM design} are generative processes because they are dependent on expert experience and training. Once the \textit{hFSM design} is completed, the modeling of the automation (states and their transitions) are determined. Thus, the processes followed are deterministic processes. \textit{hFSM classes implementation} and \textit{D-SFO (DispatcherState/Flag/Operation paradigm) implementation} are the processes that generate the codebase running in the designed system. These processes can be automated because the coded components are standardized. \textit{System testing} and \textit{hardware integration} may be semi-deterministic because of the variations of deployed environment. However, if the world modeling is optimized, these two processes should also be able to be optimized. The panel \textit{Designed System} shows the architecture of the produced system. }
    \label{fig:PCRMSDesignCycle}
\end{figure*}

\begin{figure*}[ht]
    \centering
    \includegraphics[width=\textwidth]{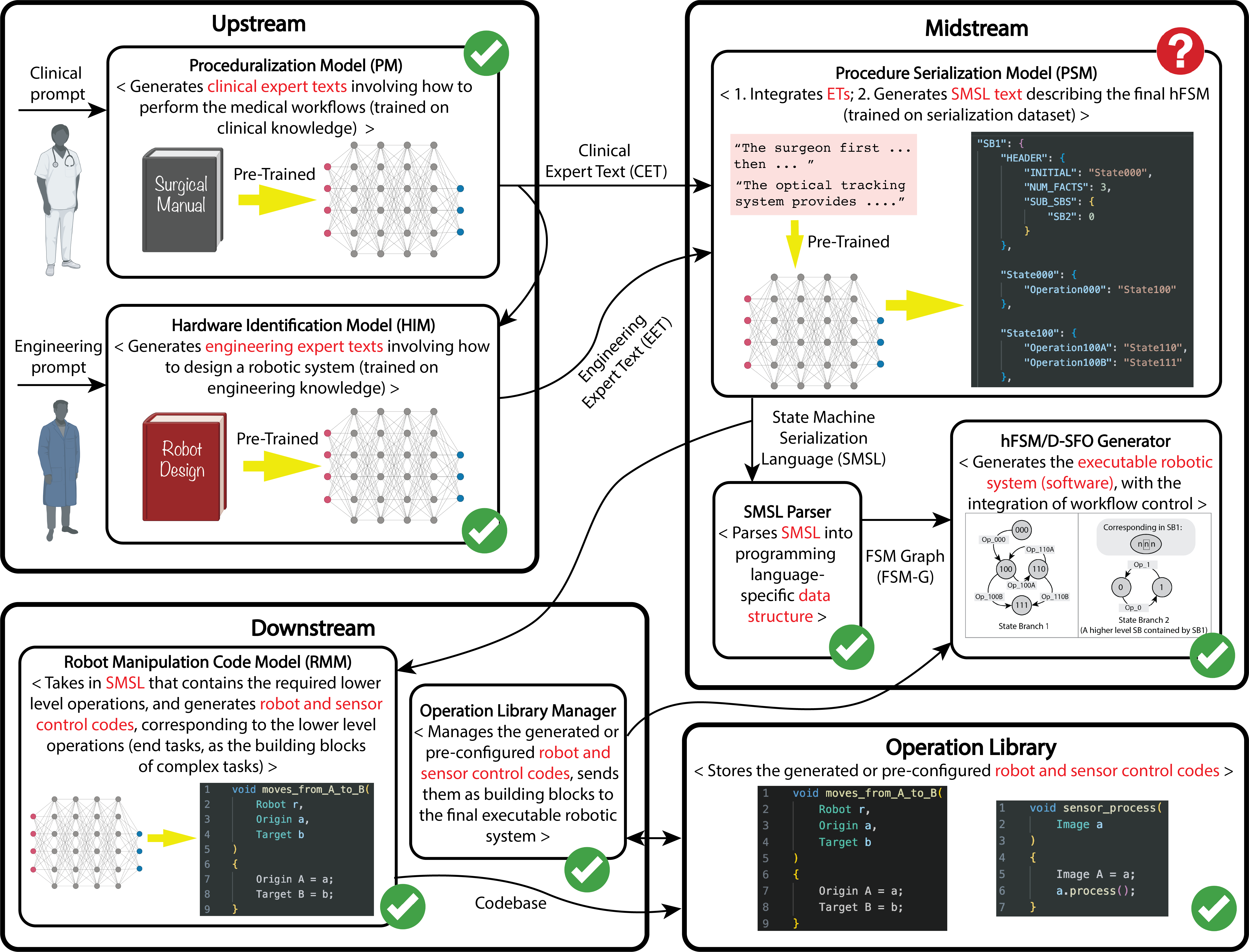}
    \caption{The architecture of the proposed paradigm, using medical robotics as an example. The modules with a green check icon are the ones that can be implemented using systems proposed in the current literature. The module with a red question mark icon is future work. The architecture approaches the intelligent robotic system in three divisions. The upstream division handles the generation of the Expert Text (ET), the midstream division involves the automatic translation of the ET to data structures in programming languages, and the downstream produces machine-executable codes for the robotic procedure, that are tailored to hardware and testing differences. }
    \label{fig:AutoArchitecture}
\end{figure*}

\section{Introduction}

Recently, the rapid development of the generative models have garnered significant attention in almost every field. Studies have shown Large Language Models (LLMs) passing or achieving high correctness in educational or professional examinations from medicine \cite{mbakwe2023chatgpt, gilson2022well, jung2023chatgpt, oztermeli2023chatgpt, fijavcko2023can, zhu2023chatgpt, ali2023performance, raimondi2023comparative, strong2023performance} to law \cite{bommarito2022gpt} and from high school \cite{wittgenstein2023tractatus, dao2023llms, dao2023evaluation, dao2023llms2} to universities \cite{bordt2023chatgpt, yeadon2024impact}. Such studies have become commonplace; the capability of the generative models and foundation models \cite{bommasani2021opportunities}, not limited to Natural Language Processing (NLP) tasks, have astonished the research community, and many predict the onset of the age of the AI. In image processing, Segment Anything Model (SAM) \cite{kirillov2023segment} has gained attention and been quickly adapted in practice \cite{liu2023samm, liu2023samme}, leading to a series of work in domain-specific applications \cite{ma2024segment, deng2023sam, zhang2023input, ke2024segment, zhao2023fast}. In video generation, Sora, a text-to-video generative model released by OpenAI, is believed to show potential capabilities of a ``world simulator'' \cite{cho2024sora, liu2024sora}. In speech generation, there is not yet a foundational model of comparable scale, but advancements are being made \cite{le2024voicebox}. This wave of recent emergence of new large scale models have demonstrated powerful capability of data understanding and generation, suggesting a promising prospect of its application in real world.  

In the field of robotics, researchers are expecting bringing automation into the next level by using generative models \cite{cui2022can, xiao2023robot}. However, the interaction with the physical world involving actuation and sensing adds more complexity to the development. These works usually focus on simpler end-tasks or simulations \cite{wang2023gensim}. Though promising, the deployment of generative models needs the integration of policy generation and online decision making, which is challenging to regulate and interpret. Due to the requirement of the interaction to the physical world, LLMs in robotics can be more risky compared to the current landmark foundation models. Moreover, reliable and governable AI is becoming a recent issue, with the latest technological advancements challenging the established norms of human society \cite{scherer2015regulating, reed2018should, wischmeyer2020regulating, erdelyi2018regulating}. In critical tasks such as medical robotics, unregulated AI and automation poses risks to safety and ethics. It is imperative to know how does robot reason and to predict what will happen in every step of the execution of a critical task. On the other hand, the benefits of using automation are obvious. Taking medical applications as an example, reliance on experts, lack of automation and scene awareness of a clinical procedure, lead to high time and money expenses in hospitals. With the development of automation systems, the society inevitably channels more resources to the research and deployment of machine intelligence assisted production. 

The core of the problem is, with some level of certainty, how to deploy LLMs in robots safely and effectively, while taking the complexity of the robotic task, the hallucination issue \cite{yao2023llm}, and the blackbox nature \cite{chao2023jailbreaking} of LLMs into consideration? In particular, can we think of ways to decompose complex tasks into simple tasks which can be generated using the current technology, while regulating the generated policy where the simple tasks are used as building blocks?  

To these important problems, we believe the proper modeling of the tasks-at-hand is crucial to regulate automation. The modeling of the tasks should contain possible configurations of a scene, and possible transitions between different scenes. The plan or policy is examined by referencing the task blueprint, then the tasks are divided and concurred by breaking them down into state transitions comprised of simple task operations. 

Finite State Machine (FSM) is well-suited for such purpose. Our previous work \cite{liu2023toward} proposes to use Hierarchical Finite State Machine (hFSM) to model the state transitions of robotic plans of a domain-specific application, translating human knowledge to machine executable instructions. In the formulation, each possible scene is modeled as a state (node) and the possible operations at each state are modeled as the edges connecting the current state and the target state. The blueprint, hFSM, can be used as the basis of the regulation. Firstly, it is a graph data structure that is easy to visualize, and secondly, it can be converted to data structures in executable robotic systems. With the formulation, we can already implement tasks using medical robots such as using da Vinci Research Kit (dVRK) \cite{kazanzides2014open} to automatically solve Hanoi Tower (Fig. \ref{fig:davincihanoi}). 

The previous work \cite{liu2023toward} also proposes foundational definitions and rules in the updates of the hFSM. We have identified deterministic steps in the design cycle, and proposed potential automation in the workflow. The gist of \cite{liu2023toward} is illustrated in Fig. \ref{fig:PCRMSDesignCycle}. Here this continued work converts those manually generative processes relying on human knowledge to automatically generative processes. We expand it to a broader roadmap, leading to the fully automatic and regulated robotic systems. This work also proposes the formalism of hFSM used in automated tasks and the regulating methods on automatically generated instructions. 

The intuition of the system in plain language is the following: The paradigm should first think about the grand goal (upstream), and come up with a plan that is connected and well-structured (midstream). This plan should be pondered upon (regulation) both before the execution (we define as ``inspection'') and during the execution (``supervision''). The smaller operations constituting the larger task, functioning as state transitions, comprise end actuation tasks. These are lower-level operations generated through training (code produced by robot foundation models, downstream). Once the regulation is completed, the plan, either the grand plan or the smaller constituents of it, can be reused. Similarly, the tiny end tasks can also be reused. As this process repeats, we can obtain a library of inspected blueprints and a library of end tasks. 

This work is to motivate future research and is not exhaustive on potential solutions. However, we believe it is safer, more controlled and more interpretable due to the nature of the techniques we propose to employ: (1) hFSM is generated preoperatively and can be inspected; (2) online supervision is possible; (3) the produced executable system follows the plan strictly. The reusablility of the inspected blueprints and validated operation library also make the paradigm standardizable. Aiming to have a paradigm where machines ``think like human'' (Section \ref{sec:upstream}), ``plan like human'' (Section \ref{sec:midstream}), ``move like human'' (Section \ref{sec:downstream}), and ``scrutinize like human'' (Section \ref{sec:regulated}), we make the following contributions:

\begin{itemize}
    \item A paradigmatic roadmap for an end-to-end, fully automated and regulated robotic system. The regulation includes both manual and automated methods for the automatically produced robotic system. Both regulating types can be done preoperatively (inspection) and intraoperatively (supervision).
    \item Formalism of the hFSM in modeling robotic tasks, and State Machine Seralization Language (SMSL), an open-sourced data language to parse structuralized expert text describing the controlled processes to a data structure in programming languages that supports the conversion to executable machine commands.
    \item A design for a Wireless Sensor Network (WSN) for state monitoring.
    \item Approaches to integrate human in the loop and a concept of human-automation continuum.
\end{itemize}

\section{Roadmap}\label{sec:roadmap}

``\textit{The world is the totality of facts, not things} \cite{wittgenstein2023tractatus}.''  The central idea of this work is to use FSM in modeling the changes of ``facts'' in a scene. In this formulation, the edges of the FSM graphs are the operations and the nodes are the states. A state contains a configuration of the facts that is in a scene. The operation between two states manipulates these facts and transitions the configuration of the facts from one to another. The management of the actuation in the robotic system are the control of the edges or operations: We model the operations that change the state of the scene to be the transitions. Taking Hanoi Tower as an example (Fig. \ref{fig:StateMachine} top panel \textit{Hanoi Tower}), each node is an abstraction of the configuration of the game pieces (disks). An operation here means the change of the position of a disk. Hence, the movements are modeled by the change of the facts in the scenes of the game. We choose FSM to model it because it is easy to visualize and to inspect, so we can regulate it after it is automatically generated. Once inspected and validated, it can be reused.

With this central idea, in this work, we articulate a roadmap using the FSM to the complete automation - an end-to-end, fully automated, and regulated robotic system. This roadmap reviews the current technology and proposes solutions. We start with the design cycle of a traditional robot-assisted procedure. From there, we integrate process control to the design cycle \cite{liu2023toward} so that the workflow is monitored and planned. Such a process-controlled design cycle is depicted in Fig. \ref{fig:PCRMSDesignCycle}. We also identify the generative and deterministic processes in the process-controlled design cycle. Then, we propose automated approaches for both the generative and deterministic processes, and eventually integrate all components from getting the human intention to producing executable systems. The system architecture of a fully automated system is shown in Fig. \ref{fig:AutoArchitecture}. Our foundational works have been done in the medical field, so we primarily use medical applications in the narratives of the roadmap.

\subsection{Overview}

A traditional design cycle of a robot-assisted procedure usually contains steps in the following. The engineering team consults with physicians with their needs and requirements in a medical scene. These needs and requirements usually starts with a demonstration of the procedure. Engineering team inspects the procedure and proposes possible point of automation. These usually include the processing of the medical images and the actuation of the medical or surgical instruments. Once identified, the engineering team consults again with the physicians and ask for an understanding of the specifications of system features and actuation goals. These idealization steps are usually iterative. With an initial agreements of the specifications, the engineering team starts the implementation of software and hardware. With prototypes, engineering and clinical teams conduct system testing together and inspect potential risks and imperfections. These steps are also iterative. The prototype is only delivered when both parties are satisfied with the test results. Then, clinical trials start with the approval of the regulatory bodies such as FDA \cite{fargen2013fda, zuckerman2011medical, van2016drugs}.

A design cycle for process-controlled system (Fig. \ref{fig:PCRMSDesignCycle}) modifies the traditional design cycle. In our previous work \cite{liu2023toward}, we have proposed an architecture of medical robotic system that integrates workflow control and modular hardware drivers, with a vision to automate not only the execution of the medical procedure, but also the design cycle of the robotic system. We have developed a paradigm that has the potential to be automatically generated. From the design cycle, we have identified that once the medical workflow has been determined, the generation of its executive instructions should be deterministic. Moreover, we have also identified that if the medical workflow is already standardized by the medical community, the result of the consultation should also be deterministic. In Fig. \ref{fig:PCRMSDesignCycle}, the design cycle contains an hFSM/D-SFO paradigm \cite{liu2023toward}, where the blueprint of the processes are defined in hFSM, and the robotic actuation codebase is in D-SFO. Once given the blueprint, the generation of the executable code is deterministic and can be automated. The downstream product, a process-controlled robotic system has been defined and implemented. 

The generative processes are those involving the generation of knowledge. Taking physician as an example, the knowledge includes proficiency in performing a surgery and understanding the movements of the surgical tools. Similarly, for an engineer, the knowledge includes how to use robots and sensors. The clinical and engineering knowledge can be structured to be an FSM design. Combining the two types of knowledge should provide viable solutions, so the FSM design is another generative process. This FSM can be complex. To simplify, we can identify steps that can be divided into sub-steps, and formulate the FSM into hFSM \cite{liu2023toward}. The iterative cycle of \textit{consultation with clinicians}, \textit{FSM design}, \textit{SB (State Branch) identification}, and \textit{hFSM design} are generative, because these all require experts' attention. However, if the task is simple, there should exist an optimal solution (an optimal robotic system), making the FSM, SB identification, and hFSM deterministic. Once the final hFSM is provided, the rest of the software and hardware implementation become deterministic: hFSM is the blueprint of the final system, so the following design procedures are already determined by the expert teams. These deterministic design processes are \textit{hFSM classes implementation}, \textit{D-SFO implementation}, and \textit{hardware integration}. The implementation of hFSM classes converts the hFSM blueprints into code. D-SFO implementation uses the hFSM classes to generate the workflow control and is integrated to the robotic system controls. Hardware integration and system testing run those generated software components. This division of generative and deterministic processes, shown in Fig. \ref{fig:PCRMSDesignCycle} (the architecture of the produced system is also shown), is the foundation of the following work.

From here, we can expand the concept and integrate generative models. This integration is with the identified generative processes and deterministic processes (Fig. \ref{fig:PCRMSDesignCycle}). A fully automated system here requires every step to be generated by AI. That is not to say the deterministic processes no long exist. Instead, the products that are expected to be produced by the deterministic processes are only delegated to AI. The delegation has constraints that conform the requirements of the deterministic processes. Thus, the products are still \textit{deterministic}. This is an important concept and we will use again in Section \ref{sec:regulated} where we discuss the regulation of the generated systems. 

With the subtle difference clarified, we start to integrate generative systems in our roadmap. Here we provide an overview. The automation that we propose involves the automation of knowledge generation and software generation. We divide the design cycle to be upstream, midstream, and downstream, based on the generated products. The overall architecture of this system is illustrated in Fig. \ref{fig:AutoArchitecture}. The upstream steps generate the text that contains expert knowledge (Expert Text, ET), from the understanding of user commands. For example: ``Generate the procedure for robotic femoroplasty''. The generative models in the upstream have been trained on both the clinical and engineering design textbooks, aiming to have the ability to generate the expert text to describe both the system and the procedure. The midstream steps first take in the ET generated from the upstream, then structures it into an hFSM representation, and finally generates the system software containing the controlled workflows and the integration of the hardware drivers. Lastly, the downstream steps generate the low-level hardware control codebase that corresponding to end-tasks. It is important to note that this roadmap and architecture design involve both current technology and speculation of the trend in the future. These components have been marked by its possibility in the current stage of the technology in Fig. \ref{fig:AutoArchitecture}. 

In the medical cases, our goal is to develop an AI assisted medical robotic system that can automatically generate process controlled medical procedures, translating clinical and engineering expert knowledge to machine executable code. The produced executables clearly define the transitions between states of the workflow, the execution of the steps, and the monitoring of the environmental conditions. The intelligent system, throughout the upstream, midstream, and downstream, includes 1. prompt-engineered LLMs to generate and serialize medical procedures from clinical ET, 2. LLMs to identify required perception, visualization and actuation hardware from engineering ET, 3. LLMs to generate the driver code for basic robot manipulations, 4. a parser to convert the serialized medical procedures in the form of natural languages into customized data structures that defines the controlled processes, and 5. a code generator to integrate the basic robot manipulation code and the process control paradigm, producing computer-interpretable and executable instructions. The hardware for the system, should be available before entering the prompts. The produced system then can be applied to robot-assisted medical procedures, such as robotic transcranial magnetic stimulation \cite{liu2023toward, liu2022inside}, knee replacement \cite{gao2020fiducial}, treatment of osteolysis \cite{sefati2020surgical}, and osteoporotic hip augmentation \cite{bakhtiarinejad2023surgical}, where researchers in the medical and engineering communities have attempted to identify and partially standardize deterministic steps, so that automated, process-controlled, and environmental-aware procedures are possible. As the community standardizes more of such procedures, the paradigm can be applied to a much broader scope, not limited to medical applications.

\subsection{Upstream - Generation of Expert Knowledge}\label{sec:upstream}

``\textit{Think like human.}'' We consider the upstream steps as the brain \cite{deng2020integrating} contemplating the grand goal to be executed. Intuitively, the process of task planning requires the education of the task, corresponding to the training of the LLM that generates the expert task. LLMs have already demonstrated capabilities that can compete human's. For example, there have been reports of LLMs passing or achieving high correctness in United States Medical Licensing Examinations (USMLE) \cite{mbakwe2023chatgpt, gilson2022well}, German State Examination in Medicine \cite{jung2023chatgpt}, Turkey Medical Specialty Exams \cite{oztermeli2023chatgpt}, American Heart Association (AHA) Basic Life Support (BLS) and Advance Cardiovascular Life Support (ACLS) exams \cite{fijavcko2023can, zhu2023chatgpt}, neurosurgery written board examinations \cite{ali2023performance}, Royal College of Ophthalmologists fellowship exams \cite{raimondi2023comparative}, free-response clinical reasoning exams \cite{strong2023performance}, high school exams on English language comprehension \cite{de2023can}, Vietnamese high school biology \cite{dao2023llms}, chemistry \cite{dao2023llms2}, physics \cite{dao2023evaluation} examinations, undergraduate algorithms and data structure exam \cite{bordt2023chatgpt}, and US legal professional license exam (the Bar Exam) \cite{bommarito2022gpt}. Most of these studies used off-the-shelf LLMs with simple prompting, making them even more promising when fine-tuned on domain-specific knowledge. 

The models in the upstreams are designed to be the bookshelves or database where the knowledge are stored, analogous to the memory in brain. Techniques such as retrieval-augmented LLMs can be applied here \cite{liu2023reta, shi2023replug, jiang2023active, nakano2021webgpt}. Such technique interfaces the LLM with external resources and retrieves documents based on user's input. The output of the retrieval-augmented LLMs is conditioned on the retrieved documents, so that the generated knowledge is relevant and valid.

However, such capability, just like understanding the mechanism of human brain \cite{petrick2020building}, is hard to measure and interpret \cite{chao2023jailbreaking}. To a large extent, they are similar to brain, and are still blackboxes simply due to the prohibitively large scale. The best that the current technology (in the ET generation in upstream) can offer is to make sure the generated knowledge is valid and constrained. Like taking a human exam, the knowledge should be tested and generated as a constrained set that is aligned to human's values and goals. In generative model research, such techniques are termed ``alignment'' \cite{wang2023aligning}. Thus, the first step of the regulation in our proposed paradigm happens in the alignment of the generative model in the upstream. More details in the regulation of the automation are provided in Section \ref{sec:regulated}.

With proper alignment (Section \ref{sec:alignment}), the Proceduralization Model (PM) and Hardware Identification Model (HIM) shown in Fig. \ref{fig:AutoArchitecture} are the upstream generative models that produce valid and correct clinical and engineering expert knowledge, respectively. Both knowledge are in the format of natural languages such as paragraphs of English. The generated expert texts can be used for other purposes, but in this paradigm, they are fed into prompt-engineered PSM (Section \ref{sec:midstream}) to generate controlled processes in the format of SMSL.

We can take a component-wise view inside the upstream here. Connected upstream and midstream (Section \ref{sec:midstream}) in the proposed intelligent medical robotic system focus on the automation of converting clinical experience to controlled and machine-executable workflows. If not integrated with generative models, the generation of the workflow, environment dependency, and procedural dependency can be replaced by: 1. Consultation with clinicians and 2. specification of system features and actuation goals. This bypasses generative models if the user choose to. On the other hand, if using the generative techniques, the PM summarizes the clinical knowledge in a condensed paragraph of the medical procedure, in the format of natural language. The HIM is used to generate engineering expert text that contains the suggested perception, visualization, and actuation hardware specifications. The HIM has an input user prompt which allows the user to specify available hardware.

All components in the upstream are possible in the current state of the generative model technology. Thus, the components here are marked as achievable using green marks in Fig. \ref{fig:AutoArchitecture}.

\subsection{Midstream - Conversion from Expert Knowledge to Machine Instructions}\label{sec:midstream}

``\textit{Plan like human.}'' Again, we can consider a human analogy here: After getting the expert experience/knowledge from the brain, human put the knowledge in the working memory \cite{angelopoulou2021working} and start to formulate steps to take in order to achieve the grand goal. This happens in the midstream of our paradigm. The ETs produced from the upstream (brain) are processed by the midstream components which generate a structured execution sequence and the modeling of the potential states of the scene. 

We first need a generative model to faithfully convert the expert knowledge to the simulation of the possible scenes. Here we restrict it to use an hFSM representation. This means the model should be able to abstract the possible states (nodes of FSM graph, or the configuration of the facts in a scene, as discussed in the beginning of Section \ref{sec:roadmap} and detailed in the formalism of FSM in Section \ref{sec:formalism}) and connect the possible states by operations (edges of FSM graph). Then, another component should take the generated FSM and convert it into a data structure which is then used in the downstream components in Section \ref{sec:downstream}.

Correspondingly, in the midstream as shown in Fig. \ref{fig:AutoArchitecture}, Procedure Serialization Model (PSM) first takes in the ET produced from PM and converts it into SMSL (Section \ref{sec:smsl}). SMSL is also in text format but is structured in a way to conform to the format requirements defined by our SMSL standards. The serialization here makes machine reading possible. Then, SMSL Parser reads SMSL, and generates a data structure (e.g. a Dictionary object in Python or a Map object in C++) containing the workflow blueprint. We define the produced data structure to be Serialized Finite State Machine Graph (FSM-G). Given the defined workflow, the driving software is deterministic \cite{liu2023toward}. Here, hFSM/D-SFO Generator takes FSM-G, producing the software with controlled workflow and integrated drivers of the hardware, shown in both Fig. \ref{fig:PCRMSDesignCycle} and \ref{fig:AutoArchitecture}.

In the D-SFO paradigm, the dispatcher and the states modules contain decision-making instructions, and the Operation Library (OL) in the downstream (Section \ref{sec:downstream}) module contains basic operations. The decision-making instructions are the converted knowledge and the basic operations are the building blocks to complex operations. Here the midstream addresses the automatic generation of machine-executable instructions. The design fills the gap toward end-to-end automation of robotic systems, providing the strict control over workflow while remaining the accuracy and stability of robots that humans cannot achieve.

In midstream, the SMSL Parser and the hFSM/D-SFO Generator are possible in the current technology. The SMSL and its parser are introduced in Section \ref{sec:smsl}, and initial work for hFSM/D-SFO has been done in our previous work \cite{liu2023toward}. So, they are marked with green marker in Fig. \ref{fig:AutoArchitecture}. However, the challenging part lies in the implementation of PSM, which is marked with a red question mark in Fig. \ref{fig:AutoArchitecture}. It is a promising future work because (1) the current LLMs already support serialization of the output in data languages \cite{wake2023chatgpt}, usually in JSON \cite{crockford2006application} format; and (2) there is evidence that more recent models have demonstrated ability to serve as ``world simulators'', and these models are usually text-to-video models \cite{chen2021geosim, blattmann2023align, wu2023tune, cho2024sora, du2024learning}.

\subsection{Downstream - Generation of Operation Library}\label{sec:downstream}

``\textit{Move like human.}'' We again use a human analogy. Humans learn the motor functions \cite{spampinato2021multiple} such as walking, running, holding, and throwing. These are also in combination with sensing functionalities \cite{qiu2022multi} such as hearing, vision, and haptics. The motor and sensing functionalities become the building blocks of more complex tasks later: A human is able to think, plan, and then execute. Such formulation of a complete cycle is top-down from the thoughts in the brain, viable plans from knowledge, and execution using the muscle and sensing functions, corresponding to our upstream, midstream, and downstream.

Thus, in downstream, we need a component that produces the execution of basic tasks. We can use a generative model Robot Manipulation Model (RMM). The generated simple tasks can be moving the robot end-effector from A to B, which causes a state change in the FSM formulation. Note the generated codebase also interfaces with sensors. To have an abundance of candidate codebase, we use a component named Operation Libraries (OL) to store previously generated or hard-coded machine instructions. These instructions comprise a validated and thoroughly tested codes, designed with a high degree of generality for reuse.

The SMSL previously produced by PSM is fed into RMM. RMM generates the OL that contains the implementations of basic manipulation functions of the hardware that are planned to be used in the procedure. The operations in the OL are the building blocks of the complex robotic procedures. They can be simple hardware actions (e.g. moving surgical tool to incision point) or human-machine interactions (e.g. waiting for visual confirmation by the surgeon). Notably, the operations in the existing OL may not be sufficient for newly added procedures, which require the development, generation, and addition of new simple tasks as building blocks of more complex robotic procedures. The dispatcher in the generated D-SFO paradigm calls the operations in the OL and controls the workflow controlled by the produced hFSM \cite{liu2023toward}. 

According to the needs, the building blocks in the OL do not have to be lower level. If a complex task is validated and can be reusable, it can also be considered to be a part of the OL. The system just need to make sure the corresponding state change is valid in such cases. Hence, the paradigm is modular, flexible and hierarchical. 

Previous works in simple robotic tasks have been done in generative robotic foundation models \cite{bommasani2021opportunities, jiang2022vima, wu2023tidybot, brohan2023rt, wang2023gensim} and more conventional reinforcement learning models \cite{ibarz2021train, hua2021learning, zhu2021deep, zhang2021reinforcement, brunke2022safe}. Therefore, the downstream components are also feasible given the current state of technology. In Fig. \ref{fig:AutoArchitecture}, these components are marked with green mark indicating that they are feasible. Nevertheless, further effort to enhance the OL is highly valuable. Through increased research within the community, the OL can incorporate more reusable, tested, and standardized execution building blocks, further simplify the paradigm and avoid the need for repetitive generation from the blackbox LLMs. We propose such actions to be standardized in the format of SMSL so that more procedures can be used from the works that already exist. 



\section{Method}

\begin{figure*}[ht]
    \centering
    \includegraphics[width=\textwidth]{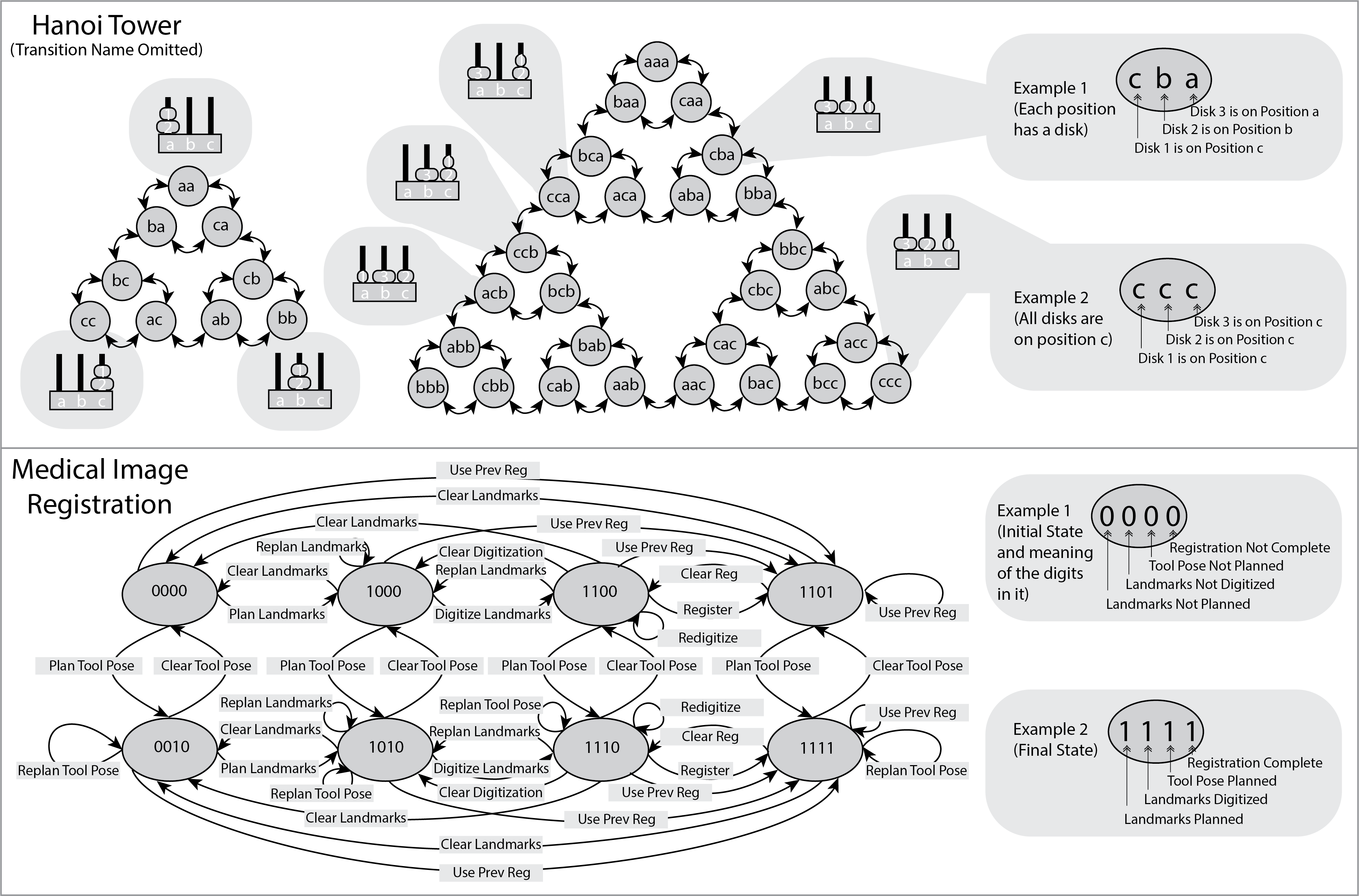}
    \caption{Example FSMs for Hanoi Tower game and for medical image registration process. The \textit{Hanoi Tower} panel shows two cases of the game: Two disks in three possible positions, and three disks in three possible positions. Fig. \ref{fig:davincihanoi} shows a simulation using the da Vinci Master Tool Manipulator (MTM) in da Vinci Research Kit (dVRK) to play Hanoi Tower automatically. The \textit{Medical Image Registration} panel is from \cite{liu2023toward}, copyright by Liu \textit{et al.} and adapted with permission. The corresponding SMSLs for both Hanoi Tower and medical image registration are listed in Appendix \ref{sec:hanoismsl} and \ref{sec:registrationsmsl}. Both examples here are not hierarchical. \cite{liu2023toward} shows the hFSM for medical image registration. For an additional simple hFSM example, Appendix \ref{sec:hfsmsmsl} lists the graph and the corresponding SMSL.}
    \label{fig:StateMachine}
\end{figure*}

\begin{figure}[ht]
    \centering
    \includegraphics[width=0.45\textwidth]{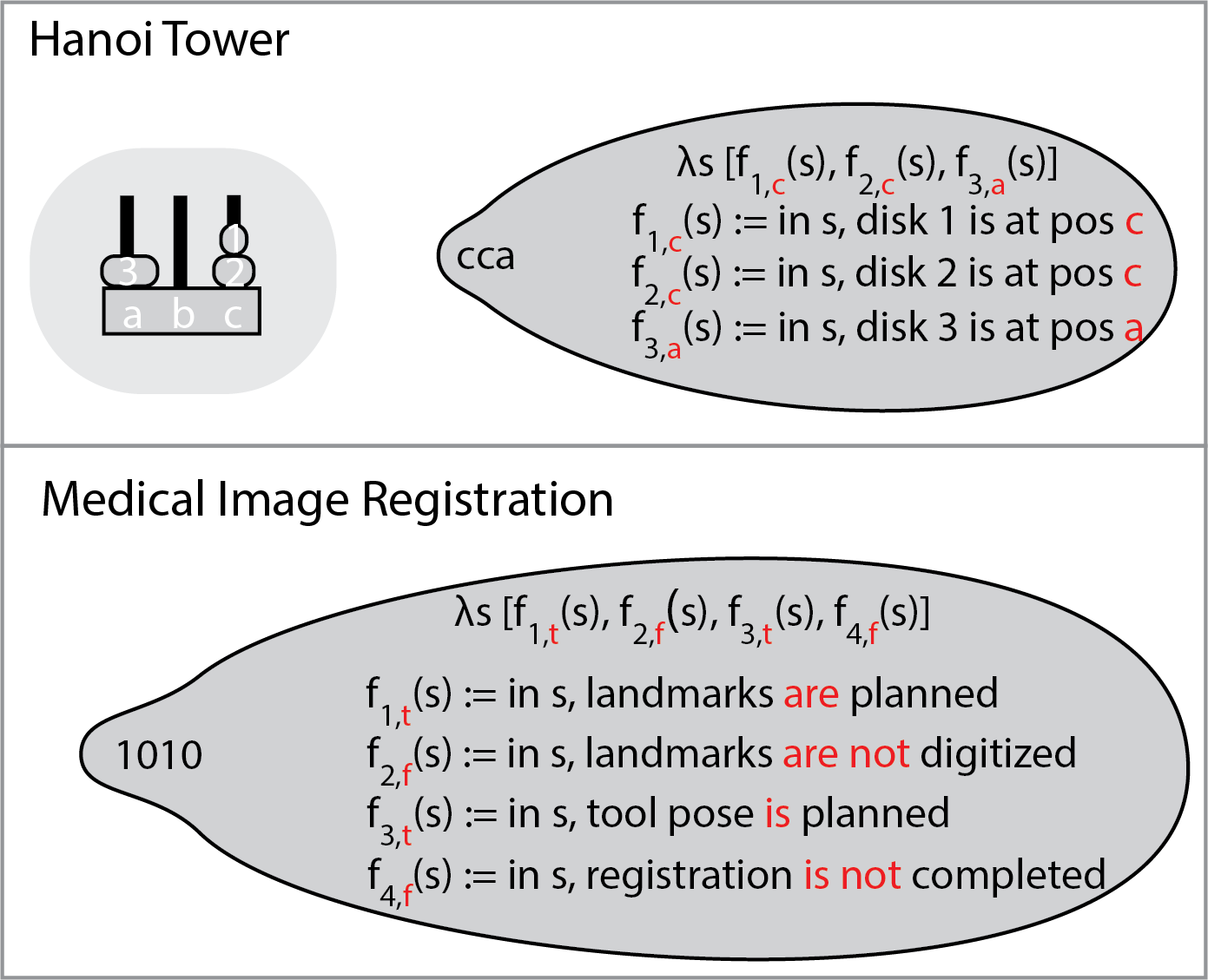}
    \caption{Modeling of the scene using lambda algebra.}
    \label{fig:lambda}
\end{figure}

\begin{figure*}[ht]
    \centering
    \includegraphics[width=0.65\textwidth]{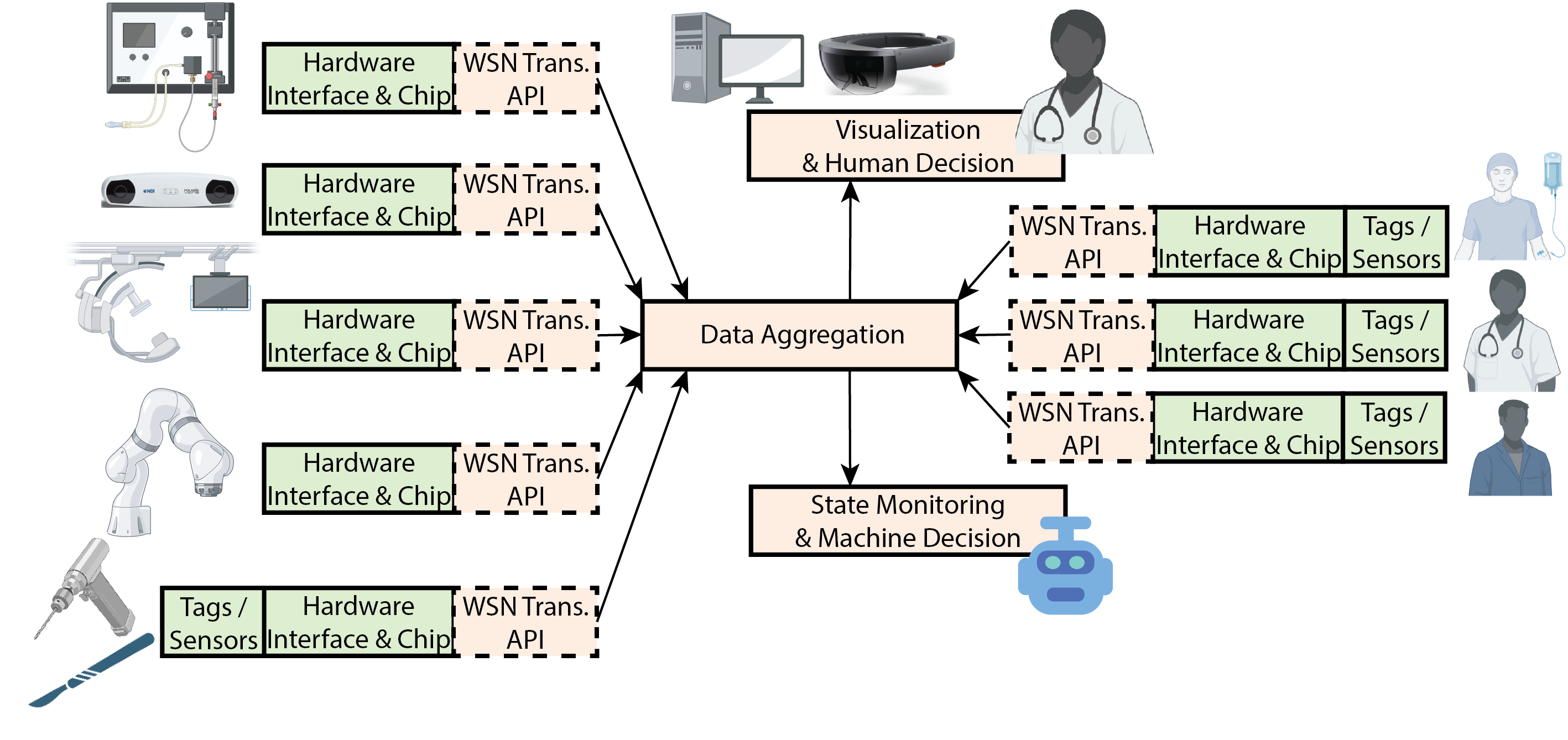}
    \caption{State monitoring using WSN and decision making. A WSN transmits the state data from all sensors to a central data aggregation module. Each sensor is either integrated to the hardware or attached as tags. The data aggregation module processes the sensor data and provides an estimation of the current state. The current state and its relevant data are then sent to the visualization module or the automated state monitoring module. Human or automation can make decisions accordingly.}
    \label{fig:WSN}
\end{figure*}

\begin{figure*}[ht]
    \centering
    \includegraphics[width=\textwidth]{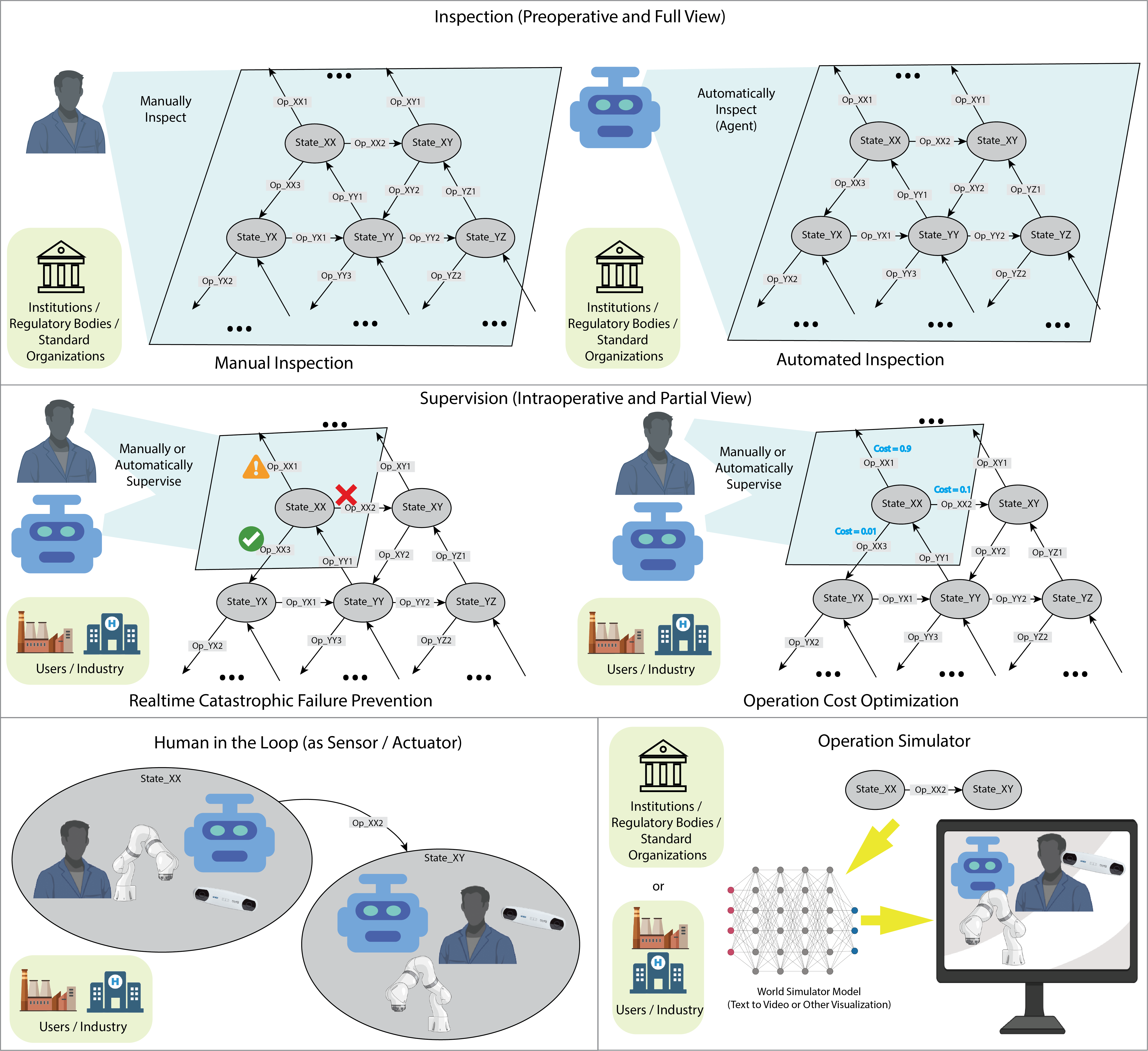}
    \caption{Regulated system and human in the loop. The regulation here is categorized into two types: Inspection and supervision. This is in addition to the alignment \cite{wang2023aligning} of LLMs, which happens at upstream and is not listed in this figure. Inspection happens preoperatively and the inspector has the full view. Inspection is conducted by institutions, regulatory bodies, or standardization organizations. Supervision happens intraoperatively and the supervisor has the partial view. The limited view is because that the supervisor is only observing the current or neighbouring states, monitoring the current scene. Both inspection and supervision can be manual, automated, or a combination of both. Human in the loop here in this bottom left panel refers when the human is acting as sensor and/or actuator in the state formulation. Human as inspector/supervisor can also be considered a type of human in the loop. During the inspection, supervision, or in-situ monitoring, a world simulator model can be employed as a predictor of the operation, which visualizes what is going to happen in the state transition. }
    \label{fig:regulated}
\end{figure*}

The overarching goal of our work is to conceptualize an intelligent system capable of automatically generating process-controlled workflow and machine executable instructions tailored to available hardware. In a medical robotics case, the system includes 1. Proceduralization Model (PM) to generate clinical Expert Text (ET), 2. Hardware Identification Model (HIM) to generate engineering ET with available hardware, 3. Procedure Serialization Model (PSM) to combine the ETs from clinical and engineering domains and generate serialized SMSL, 4. Robot Manipulation Model (RMM) to generate basic robot manipulation code as building blocks to complex tasks, 5. SMSL and its parser to convert ET in the format of natural languages to FSM-Graphs (FSM-Gs) that can be used in common programming languages, and 6. hFSM/D-SFO generator to combine the process control data and the robot manipulation code to generate the executable robotic system.

This and the next section is divided into subsections to introduce the theory, practical implementation, and considerations of some notable innovations we are proposing. These include the formalism of the FSM abstract in the paradigm, a new data language SMSL for serializing FSMs, a Wireless Sensor Network (WSB) used to monitor state changes, the regulation of the processes involved in the end-to-end automation, failure mode handling utilizing the graph data structure, and human in the loop.

\subsection{Formalism of State Modeling}\label{sec:formalism}

The core of the midstream in the paradigm is to model the scene in a distinct and structured way. Considering our technical approach, hFSM, we model the scenes and the relevant factors as a combination of ``facts''. Intuitively, each state (a node in the FSM graph) contains a collection of facts, and each operation (an edge in the FSM graph) induces a change in the facts in the scene. These facts and their manipulations are converted from the expert text produced in the upstream. Thus, the extraction of the meaning (semantics) of natural languages need to happen in the process. A common way to model semantics of sentences is to use lambda calculus \cite{morris1968lambda, cardone2006history, wong2007learning}. Shown in \cite{liu2023toward}, the existing facts in a scene can be annotated as a binary true or false. In this way, a combination of facts are configured to be a binary string. However, they can also be non-binary, depending on the specific needs of the description of the state. The Hanoi Tower (top panel Fig. \ref{fig:StateMachine}) uses a non-binary annotation (instead of 0s and 1s, it uses a, b, c). These facts can be the returned results of sensors. The combination of the identification of all the facts-of-interest becomes scene identification.

\subsubsection{State Representation}

\cite{liu2023toward} proposed definitions and rules to use hFSM in workflow control, but does not formulate the mathematical model. Here we build on the previous work and provide the formalism of the concept. We can use lambda calculus to model a state. We consider a simple case where we suppose in the scene $s$, facts $f_1$ and $f_2$ are true. We can write:

\begin{center}
    \textit{There exist a state $s$, where $f_1$ and $f_2$ are true.}
\end{center}

Or, in a more compact way:

\begin{equation}
    \exists s \textnormal{ } f_1(s) \textnormal{ AND } f_2(s)
\end{equation}

Equivalently:

\begin{equation}
    exists(\lambda s \textnormal{ } f_1(s), \lambda s \textnormal{ } f_2(s))
\end{equation}

Thus, using lambda calculus, we can formulate a state check to any scene of interest. This state check is to test whether the state is activated. We can then define the State Check Function (SCF) to be:

\begin{equation}
    \lambda s \textnormal{ } [f_1(s),f_2(s)]
\end{equation}
where $s$ is the scene variable. The SCF returns a boolean. More generally, for a state definition with multiple facts, we can define:

\begin{equation}\label{eq:statecheck}
    C_{state}(s):= \lambda s \textnormal{ } [\Lambda_{i=1}^N \textnormal{ } f_i(s)]
\end{equation}

\subsubsection{Sensor Representation}

With the established annotation, sensor checks can be formulated. In this representation, each $f_i$ is obtained by a sensor or a combination of the sensors:

\begin{equation}\label{eq:sensorcheck}
\begin{split}
    f_i&:=C_{fact}(p,s)\\
    C_{fact}(p,s)&:= \lambda p \textnormal{ } \lambda s \textnormal{ } F(sense(p, s))
\end{split}
\end{equation}
where $p$ is a variable handling a sensor or a combination of sensors, and $F$ is the predicate describing the desired sensor results. Again, $f_i$ returns a boolean. These definitions can be seen in the examples provided in Fig. \ref{fig:lambda}.

\subsubsection{Operation Representation}

The representation of the operations (state transitions) is to pass a scene variable that represent the current scene. So, each edge is a change of the SCF. 

\subsubsection{Hierarchical State Representation}

To simplify an FSM, we can use hFSM \cite{liu2023toward, sklyarov1999hierarchical}. The formulation of a sub-State Branch (termed ``higher level SB'' or ``$N+1$ SB'' in \cite{liu2023toward}) is to describe an $f_i$ using a state check. The state check is the same as in Eq. \ref{eq:statecheck}.

\begin{equation}
\begin{split}
    f_i&:=C_{state}(s_{sub})\\
    C_{state}(s_{sub})&:= \lambda s_{sub} \textnormal{ } [\Lambda_{i=1}^N \textnormal{ } f_{sub,i}(s_{sub})]
\end{split}
\end{equation}

Thus, formally, the hFSM formulation of the workflow is: 1. A multidigraph; 2. each edge is a permitted operation that changes the SCF; 3. each node is a SCF as formulated above; 4. each fact in the state can be expanded as a sub-SB that contains an activating state.

\subsection{State Machine Serialization Language}\label{sec:smsl}

The outputs from the upstream are paragraphs of ETs that are in the format of natural languages. The ETs can be read by midstream models as prompts. Already discussed in Section \ref{sec:midstream}, a PSM converts the ETs to a structured data that can be both read by machine and human. This structured data is the input of the hFSM/D-SFO Generator. So, we need to define this intermediate data language.

Here in this work we introduce SMSL as a new data serialization language readable by both human and machine. In the implementation, the expert knowledge is converted to serialized FSM-G data that can be stored and used by programming languages. It defines the format of the data storage and the text file syntax. SMSL aims to achieve similar manifestation as Yet Another Markup Language (YAML) \cite{ben2009yaml}, Extensible Markup Language (XML) \cite{bourret1999xml}, and JSON \cite{crockford2006application}, but focuses on the serialization of hFSM definitions. As a part of the SMSL framework, the SMSL Parser reads SMSL text and converts the stored hFSM definition to data structures in common programming languages. The code repository of SMSL is available at https://smsl.dev .

SMSL is a data language that can be parsed into a graph data structure. The examples are provided in Appendix \ref{sec:hanoismsl} (Hanoi Tower), \ref{sec:registrationsmsl} (medical image registration), and \ref{sec:hfsmsmsl} (hFSM). The corresponding graphs are shown in Fig. \ref{fig:StateMachine} top panel, Fig. \ref{fig:StateMachine} bottom panel, and Fig. \ref{fig:hFSM}, respectively. An SMSL file contains State Branches (SB). Each SB is an FSM and contains multiple possible states. Correspondingly, these states are the nodes in the graph representations. Each state has possible operations, which correspond to the edges of the graph representation. For example, in the Hanoi Tower SMSL (Appendix \ref{sec:hanoismsl}), ``State\_aaa'' represents the state where all disks are at position \textit{a}, and it contains two possible operations: ``Op\_1b'', moving disk \textit{1} to position \textit{b}, and ``Op\_1c'', moving disk \textit{1} to position \textit{c}. In such way, a text file can represent the complete graph of an FSM. Each SB may also contain a ``HEADER'' field. The header field is mainly used for hFSMs, where ``SUB\_SBS'' contains the information of the sub-SBs. This field connects the sub-SBs to the base-SB by identifying which ``fact'' in base-SB is the corresponding sub-SB. For example, ``SB2 : 1'' means ``SB2'' (sub-SB or higher level SB) corresponds to the 2nd (``1''+1, zero-based numbering) fact in SB1 (base-SB or lower level SB). SB2 has an activating state, which means if the scene is at that state, the 2nd fact of the corresponding lower level SB1 is also activated. 

\subsection{State Monitoring Wireless Sensor Network}

A successful automation should be ``aware'' of its surroundings. The modeling of the states is based on such situational awareness where the states of the scene is constantly monitored. In our formulation of the automation, each state contains a set of ``facts'' and has been formulated as a state check function in Equation \ref{eq:statecheck}. These ``facts'' of the scene, are check by sensors, as formulated in Equation \ref{eq:sensorcheck}. 

For a scene that contains multi-modal sensors, a popular implementation for data collection is to use a Wireless Sensor Network (WSN) \cite{raghavendra2006wireless, kandris2020applications}. The benefits of WSN in many applications have been extensively explored in the literature regarding its scalability \cite{pakzad2008design, dogra2021essence, calderwood2020low},  energy efficiency / passivity \cite{juang2002energy, amutha2020wsn, nakas2020energy, yang2022hybrid}, latency \cite{scanzio2020wireless}, and adaptation \cite{zivkovic2021enhanced}. These advancements of WSN and their extension have become a dedicated field, namely, the Internet of Things (IoT) \cite{laghari2021review}.

Here we again use medical automation as an example. Previous studies have investigated the types of situation awareness in an operating room (OR) \cite{kranzfelder2011new}. These types include: Functional state of peripheral devices and systems, recognition of surgical instruments, behavior of the team, and emotion analysis of the persons involved. Similar research can be seen in both human factors \cite{graafland2015training} and non-human factors \cite{gao2006vital}. In an automated medical robotic system, state monitoring can be designed in the way demonstrated in Fig. \ref{fig:WSN}. In similar cases, the functional state of the peripheral devices or hardware can be extended to all robotic and intraoperative imaging devices, which is lacking in the literature. The data from active actuation and sensing units, including robots, medical devices, and intraoperative imaging instruments, can be collected in real time using an unified WSN transmission interface similar or with modifications to the existing WSNs in various applications in \cite{juang2002energy, mottola2010not, hodge2014wireless, li2022application, afanasov2020battery}. Realtime data collection from passive instrument, most commonly the localization data, can use tagging systems like \cite{zhao2020nfc+} (lower accuracy requirement) or active localization systems in \cite{bianchi2019localization} (higher accuracy requirement). Human factors can be monitored using vision, speech, and localization-based sensors.

The state monitoring WSN collects all modalities of the data and aggregates them. The data aggregation module in Fig. \ref{fig:WSN} is the state and sensor checks as formulated in Equation \ref{eq:statecheck} and \ref{eq:sensorcheck}. The state monitoring is then tied back to the product in the midstream where the machine instructions (decision making) happen according to the current state. This decision making can be human and/or automation, discussed in Section \ref{sec:humanintheloop} and \ref{sec:humanautocontinuum}.

\subsection{Regulated Workflow}\label{sec:regulated}

``\textit{Scrutinize like human.}'' A missing piece in the current technology is that the AI is not sufficiently regulated and monitored. There has not been an agreed-upon consensus in how to carry out such regulation endeavour \cite{scherer2015regulating, reed2018should, wischmeyer2020regulating, erdelyi2018regulating}.  The regulation is challenging due to the black-box nature of the AI and the prohibitively large scale of the existing models. The interpretability of machine learning is a fairly recent topic \cite{molnar2020interpretable, du2019techniques, rudin2022interpretable} and the challenges are more significant in generative AI \cite{nalisnick2018deep, park2024explaining}. Such unclear interpretations is complicated by the size of the generative models: A well-known example is the hallucinations in LLM \cite{yao2023llm} where outputs can be fabricated and non-existent facts. In light of this, we propose the regulation of the automation on the products of the AI, not the execution of the AI. In other words, as long as we make sure the products produced by AI is trustable, we can use the products. Similar to the inspection and review processes applied to man-made products, humans are prone to errors more frequently. However, society has developed mechanisms to detect and address these errors. To this, we design three layers of safety, including alignment, inspection, and supervision. The process of these layers in our paradigm are analogous to ``scrutinize while learning'', ``scrutinize after planning'', and ``scrutinize while executing''.

\subsubsection{Alignment}\label{sec:alignment}

``\textit{Scrutinize while learning.}'' The first layer of security is done by alignment, where the generation of the knowledge in the upstream is steered to be correct. The alignment of LLMs is to ensure the models interact with humans in the useful and unharmful way \cite{wang2023aligning, wolf2023fundamental}. Concerns have been circulating in using generative models. The models can exhibit fake information \cite{wolf2023fundamental, lin2021truthfulqa, weidinger2022taxonomy}, social biases \cite{hutchinson2020social}, encouragement of questionable behaviors \cite{roose2023conversation, wang2023aligning}. Current work in alignment in LLMs are focused on the alignment to the value of the human so that the information provided by the models are truthful and correct. Here in our paradigm, the upstream models are considered to be the knowledge generators, so the ETs are required to be useful and truthful. In particular, the models in the upstream can serve as a database of the knowledge. The knowledge that the models are trained on should be constrained, and the trained models should be aligned using injected prompts or reinforcement learning from human feedback (RLHF) \cite{wolf2023fundamental}.

However, the alignment of the models is not guaranteed to be satisfactory. At the same time, the alignment process itself is also not deterministic, which is difficult to be regulated. Thus, we add the inspection and the supervision of the system, and both happen \textit{after} the knowledge generation.

\subsubsection{Inspection}

``\textit{Scrutinize after planning.}'' The plans are generated by the midstream, and can be visualized using graphs. The inspection process is the same as proposing the plans in the blueprints and regulatory bodies inspecting the blueprints. The inspector has the full view of these blueprints, as shown in Fig. \ref{fig:regulated} top panel. The advantage of generating policy in the format of SMSL is that the policy can be inspected: The generated results of SMSL is in the format of text, and the data language allows the parsing of the text data into graphical representations. The graphs (as in data structure), can be visualized using graph visualization toolkits. After each inspection, the plans can be reused. 

The inspection process can be done manually or potentially automatically. However, for automatic inspection, additional models are required to be able to understand and identify risks. These automatic inspection models should also be regulated (inspected). Current state of the technology has not investigated the feasibility of risk estimation, but it is a promising field. There is evidence that some text-to-video LLMs can serve as ``world simulators'' \cite{cho2024sora, du2024learning}. The paradigm can use the world simulation feature to predict the outcome of the operation (state transition), as shown in Fig. \ref{fig:regulated} bottom right panel.

It is worth to note that the process of generating the policy can be online (in realtime). In this case, the graph generation of the FSM-G is in real time. Here, at a node, the PSM searches for possible output based on the current state (branching out in the graph). However, the online generation can be risky because the policy is not fully regulated. The online generation should only be used based on the needs and the speed of supervision.

\subsubsection{Supervision}

``\textit{Scrutinize while executing.}'' Similarly to inspection, shown in the middle panel of Fig. \ref{fig:regulated}, during the execution of robotic tasks, the paradigm adds the supervision layer where the state transitions are monitored. The supervision process happens using the partial view of the blueprint: Only the current state and its adjacent states and operations are visible to the supervisor. The supervisor is able to visualize and estimate the risk while the execution is online, and can take over if necessary. The monitoring can be on the physical scene, the serialized SMSL, or visualized FSM-G.

\section{Discussion}
The methods can be extended to some valuable extensions and concepts. Here we discuss some of the features that are beneficial including how to modularize the processes of the regulation and the libraries of the operations, how to integrate failure handling in the FSM, and what roles human would play in the paradigm. 
 
\subsection{Modular State and Operation Library}

\subsubsection{Reuse Inspected SMSL}
The state graphs can be reused. Common processes can be encapsulated and inspected separately. Once inspected and tested, the process (an FSM-G) can be used in other applications. Standard inspection can be established and these inspections can be delegated to specialized organizations in the industry. In this way, the generation of the expert knowledge and the conversion from the knowledge to SMSLs only need to be carried out once or few times. 

\subsubsection{Reuse Generated Operation Libraries}
Similarly, the generation of the OLs that contain the codebase for the building block operations can be reused. Again, once generated, inspected, and tested, the encapsulated libraries are stored and standardized.

\subsection{Failure Mode Handling}

\subsubsection{Expected Failure}
The failure mode handling is part of the paradigm. In the ETs, expert handling of errors should be part of the knowledge. For example, the doctors are trained to know the emergencies from the medical knowledge. In this case, a state may contain ``facts'' that are the undesirable failures. Based on the generated blueprints, paths should be able to be planned toward correcting the failures and eventually achieve a failure-free state.

\subsubsection{Realtime Catastrophic Failure Prevention}
With the monitoring of the states, if an already-planned operation is temporarily invalid, it can be tagged to be risky, so the routing of the automation would bypass the operation (Fig. \ref{fig:regulated} middle panel left side, considering the edge to be pruned). 

\subsubsection{Operation Cost Optimization}
Cost functions can be associated to each operation (edge). In such cases, the routing of the automation will search for the shortest path toward the goal state. This cost function can be adaptive, depending on the current state. Thus, the automation problem in the scope of FSM-G becomes an optimization problem.

\subsection{Human in the Loop}\label{sec:humanintheloop}

Human in the loop in the proposed paradigm can be in various formats. Most directly, the inspection of the plan involves with human. However, here we only consider the human in the loop as in the human in the execution time of the produced executable system. 

\subsubsection{Human as Realtime Supervisor}

As shown in the middle panel of Fig. \ref{fig:regulated}, the human involvement can be the supervision of the realtime execution. In this case, the human is not a part of the execution. The human here only serves as the supervisor to monitor the automated scene and would take over if risks are identified. Human in this case is not directly in the loop. As depicted in the middle panel of Fig. \ref{fig:regulated}, the human (or automated agent) is outside of the states.

\subsubsection{Human as Actuator / Sensor}

Another case of human in the loop is human as actuators and sensors. In this case, the human or the automated agent is inside the states. The state of the human and the automated agent is a part of the overall FSM, and is contained to be ``facts'' in the scene. This happens more commonly when there exists limitation of the automation, so some tasks are still manual. In the automated scene, human operations and robotic operations are combined. 

\begin{figure}
    \centering
    \includegraphics[width=0.5\textwidth]{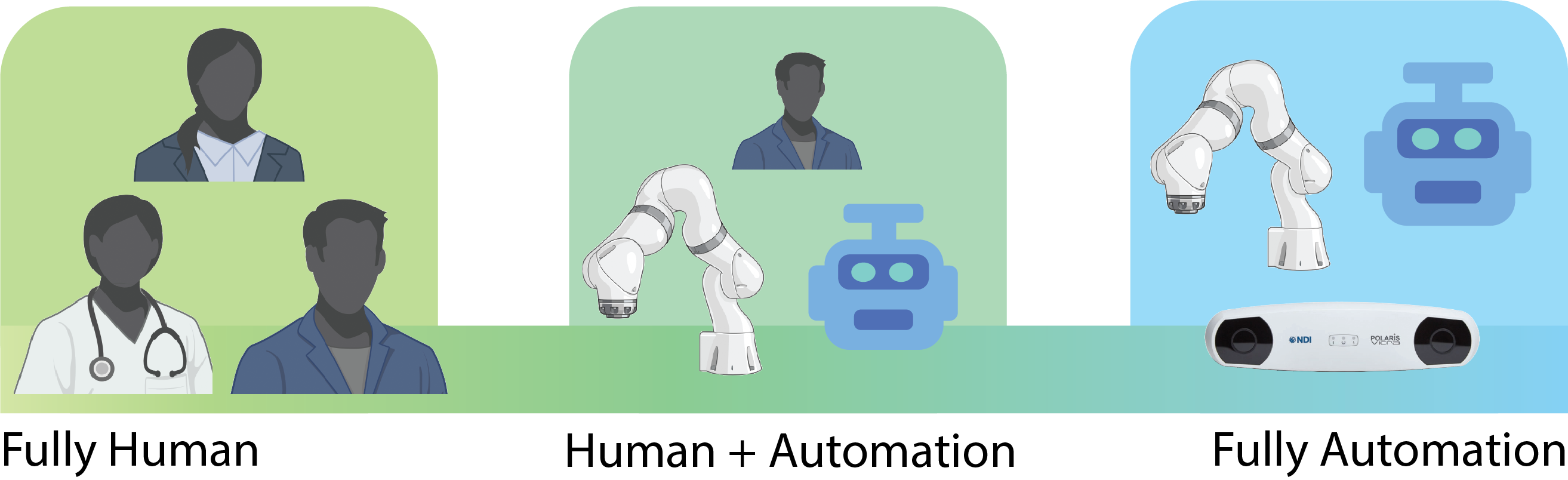}
    \caption{Human in the loop: Human-automation continuum. Our formulation of the roadmap considers the roles of human and automation to be a continuum. From left to right, as the color changes in the spectrum (green: human, blue: automation), a scene can be composed of only human to only automation (actuator/sensor/agent). This scene can be the inspection, surpervision, or state operation/actuation (Fig. \ref{fig:regulated}). The location of the spectrum of a implemented system depends on the need and the availability of the technology. }
    \label{fig:continuum}
\end{figure}

\subsection{Human-Automation Continuum}\label{sec:humanautocontinuum}
The development of generative models makes the boundary between generative processes and deterministic processes vague. It is most likely that human will still play an imperative role in the future. An integration of human to the automation paradigms will likely be the norm. Here we have examined a potential framework of fully automated and regulated robotic system. Now we propose a concept of human-automation continuum.

Already discussed, humans can serve as actuators and sensors as parts of the system. Therefore, a combination of the human and automation exist as shown in Fig. \ref{fig:continuum}, where the continuum contains a spectrum from fully human to fully automation. The spectrum is likely to stay in the middle for some time because replacing human from production is a process that takes a long time. This is true similarly in the continuum in the regulation process. Both inspection and supervision can be potentially performed by human, automation or the combination, depending on the capability of the automated agent. 

The continuum in workflow generation can be a promising research area. It may exist because of the limitation of the ``world simulator'' models (PSM in midstream). Some simple scenes such as Hanoi Tower is easy to exhaustively iterate and generate with the pre-defined game rules. However, for more practical and complex scenes, it is obviously challenging. In this case, the SMSL data generation can be done manually or in a combined way of manual and automated methods. Thus, the continuum exist.

\section{Conclusion}

Achieving end-to-end automation in robotic systems is a complex system engineering topic, involving various disciplines from upstream to downstream. The emerging generative technology has the potential to accelerate the development but is facing challenges in practical application, interpretation and regulation. In this work, we provide a roadmap that can potentially achieve a paradigm for fully automated and regulated robotic systems. We divide the paradigm into upstream, midstream, and downstream, corresponding to the parts to generate the expert knowledge, convert the knowledge into controlled workflow, and generate the building block codebase for the hardware. We propose SMSL, a text-based FSM serialization data language, to assist the interfacing between the generated expert knowledge and the executable machine instructions. We also provide the formalism of the controlled FSM. Moreover, regulation methods are provided throughout the streams. Finally, we also articulate the concept of the human-automation continuum. 

\section*{Acknowledgement}

Some of the figures in this work contain logos and graphics created with BioRender.com: Engineer, doctor, employee, book, Kuka iiwa robot, desktop computer, CArm, drilling tool, surgical knife, patient, anesthesia machine, factory, and hospital.

\bibliographystyle{IEEEtran}
\bibliography{bib} 

\newpage
\begin{appendices}
\section{SMSL Example 1: Hanoi Tower}\label{sec:hanoismsl}
The FSM of the following SMSL is shown in Fig. \ref{fig:StateMachine}. Naming convention used in this example is consistent to Figure \ref{fig:StateMachine}. For states, the names are ``State\_'' followed by the position of disk \textit{1}, position of disk \textit{2}, and position of disk \textit{3}. For example, ``State\_aaa'' means the disk \textit{1} is at position \textit{a}, disk \textit{2} is at position \textit{a}, and disk \textit{3} is at position \textit{a}. For operations, the names are ``Op\_'' followed by the \textit{disk to be moved}, then followed by the position the disk to be moved to. For example: ``Op\_1b'' means ``move disk \textit{1} to position \textit{b}''.

\begin{lstlisting}[language=json, linewidth=8.5cm]
{
    "SB1": {
        "HEADER": {},
        "State_aaa": {
            "Op_1b": "State_baa",
            "Op_1c": "State_caa"
        },
        "State_baa": {
            "Op_1a": "State_aaa",
            "Op_1c": "State_caa",
            "Op_2c": "State_bca"
        },
        "State_caa": {
            "Op_1a": "State_aaa",
            "Op_1b": "State_baa",
            "Op_2b": "State_cba"
        },
        "State_bca": {
            "Op_1c": "State_cca",
            "Op_1a": "State_aca",
            "Op_2a": "State_baa"
        },
        "State_cca": {
            "Op_1b": "State_bca",
            "Op_1a": "State_aca",
            "Op_3b": "State_ccb"
        },
        "State_aca": {
            "Op_1b": "State_bca",
            "Op_1c": "State_cca",
            "Op_2b": "State_aba"
        },
        "State_cba": {
            "Op_1a": "State_aba",
            "Op_1b": "State_bba",
            "Op_2a": "State_caa"
        },
        "State_aba": {
            "Op_1c": "State_cba",
            "Op_1b": "State_bba",
            "Op_2c": "State_aca"
        },
        "State_bba": {
            "Op_1c": "State_cba",
            "Op_1a": "State_aba",
            "Op_3c": "State_bbc"
        },
        "State_ccb": {
            "Op_3a": "State_cca",
            "Op_1a": "State_acb",
            "Op_1b": "State_bcb"
        },
        "State_acb": {
            "Op_1c": "State_ccb",
            "Op_1b": "State_bcb",
            "Op_2b": "State_abb"
        },
        "State_bcb": {
            "Op_1c": "State_ccb",
            "Op_1a": "State_acb",
            "Op_2a": "State_bab"
        },
        "State_abb": {
            "Op_2c": "State_acb",
            "Op_1b": "State_bbb",
            "Op_1c": "State_cbb"
        },
        "State_bbb": {
            "Op_1a": "State_abb",
            "Op_1c": "State_cbb"
        },
        "State_cbb": {
            "Op_1c": "State_bbb",
            "Op_1a": "State_abb",
            "Op_2a": "State_cab"
        },
        "State_bab": {
            "Op_2c": "State_bcb",
            "Op_1c": "State_cab",
            "Op_1a": "State_aab"
        },
        "State_cab": {
            "Op_2b": "State_cbb",
            "Op_1b": "State_bab",
            "Op_1a": "State_aab"
        },
        "State_aab": {
            "Op_1b": "State_bab",
            "Op_1c": "State_cab",
            "Op_3c": "State_aac"
        },
        "State_bbc": {
            "Op_3a": "State_bba",
            "Op_1c": "State_cbc",
            "Op_1a": "State_abc"
        },
        "State_cbc": {
            "Op_1b": "State_bbc",
            "Op_1a": "State_abc",
            "Op_2b": "State_cac"
        },
        "State_abc": {
            "Op_1b": "State_bbc",
            "Op_1c": "State_cbc",
            "Op_2c": "State_acc"
        },
        "State_cac": {
            "Op_2b": "State_cbc",
            "Op_1a": "State_aac",
            "Op_1b": "State_bac"
        },
        "State_aac": {
            "Op_3b": "State_aab",
            "Op_1c": "State_cac",
            "Op_1b": "State_bac"
        },
        "State_bac": {
            "Op_1c": "State_cac",
            "Op_1a": "State_aac",
            "Op_2c": "State_bcc"
        },
        "State_acc": {
            "Op_2b": "State_abc",
            "Op_1b": "State_bcc",
            "Op_1c": "State_ccc"
        },
        "State_bcc": {
            "Op_1c": "State_ccc",
            "Op_2a": "State_bac",
            "Op_1a": "State_acc"
        },
        "State_ccc": {
            "Op_1a": "State_acc",
            "Op_1b": "State_bcc"
        }
    }
}
\end{lstlisting}

\newpage
\section{SMSL Example 2: Medical Image Registration}\label{sec:registrationsmsl}

The FSM of the following SMSL is shown in Fig. \ref{fig:StateMachine}.

\begin{lstlisting}[language=json, linewidth=8.5cm]
{
    "REGISTRATION": {
        "HEADER": {},
        "State_0000": {
            "Op_PlanLandmarks": "State_1000",
            "Op_PlanToolPose": "State_0010",
            "Op_UsePrevReg": "State_1101"
        },
        "State_0010": {
            "Op_ClearToolPose": "State_0000",
            "Op_ReplanToolPose": "State_0010",
            "Op_PlanLandmarks": "State_1010",
            "Op_UsePrevReg": "State_1111"
        },
        "State_1000": {
            "Op_ClearLandmarks": "State_0000",
            "Op_ReplanLandmarks": "State_1000",
            "Op_UsePrevReg": "State_1101",
            "Op_DigitizeLandmarks": "State_1100",
            "Op_PlanToolPose": "State_1010"
        },
        "State_1010": {
            "Op_ClearLandmarks": "State_0010",
            "Op_ReplanLandmarks": "State_1010",
            "Op_ClearToolPose": "State_1000",
            "Op_DigitizeLandmarks": "State_1110",
            "Op_ReplanToolPose": "State_1010",
            "Op_UsePrevReg": "State_1111"
        },
        "State_1100": {
            "Op_ClearDigitization": "State_1000",
            "Op_ReplanLandmarks": "State_1000",
            "Op_ClearLandmarks": "State_0000",
            "Op_Redigitize": "State_1100",
            "Op_Register": "State_1101",
            "Op_UsePrevReg": "State_1101",
            "Op_PlanToolPose": "State_1110"
        },
        "State_1110": {
            "Op_ClearLandmarks": "State_0010",
            "Op_ReplanLandmarks": "State_1010",
            "Op_ClearToolPose": "State_1100",
            "Op_ReplanToolPose": "State_1110",
            "Op_Redigitize": "State_1110",
            "Op_ClearDigitization": "State_1010",
            "Op_UsePrevReg": "State_1111",
            "Op_Register": "State_1111"
        },
        "State_1101": {
            "Op_ClearLandmarks": "State_0000",
            "Op_ClearReg": "State_1100",
            "Op_UsePrevReg": "State_1101",
            "Op_PlanToolPose": "State_1111"
        },
        "State_1111": {
            "Op_ClearReg": "State_1110",
            "Op_ClearToolPose": "State_1101",
            "Op_UsePrevReg": "State_1111",
            "Op_ReplanToolPose": "State_1111",
            "Op_ClearLandmarks": "State_0010"
        }
    }
}
\end{lstlisting}

\newpage
\section{SMSL Example 3: Simple Hierarchical FSM}\label{sec:hfsmsmsl}

\begin{figure}[h]
    \centering
    \includegraphics[width=0.45\textwidth]{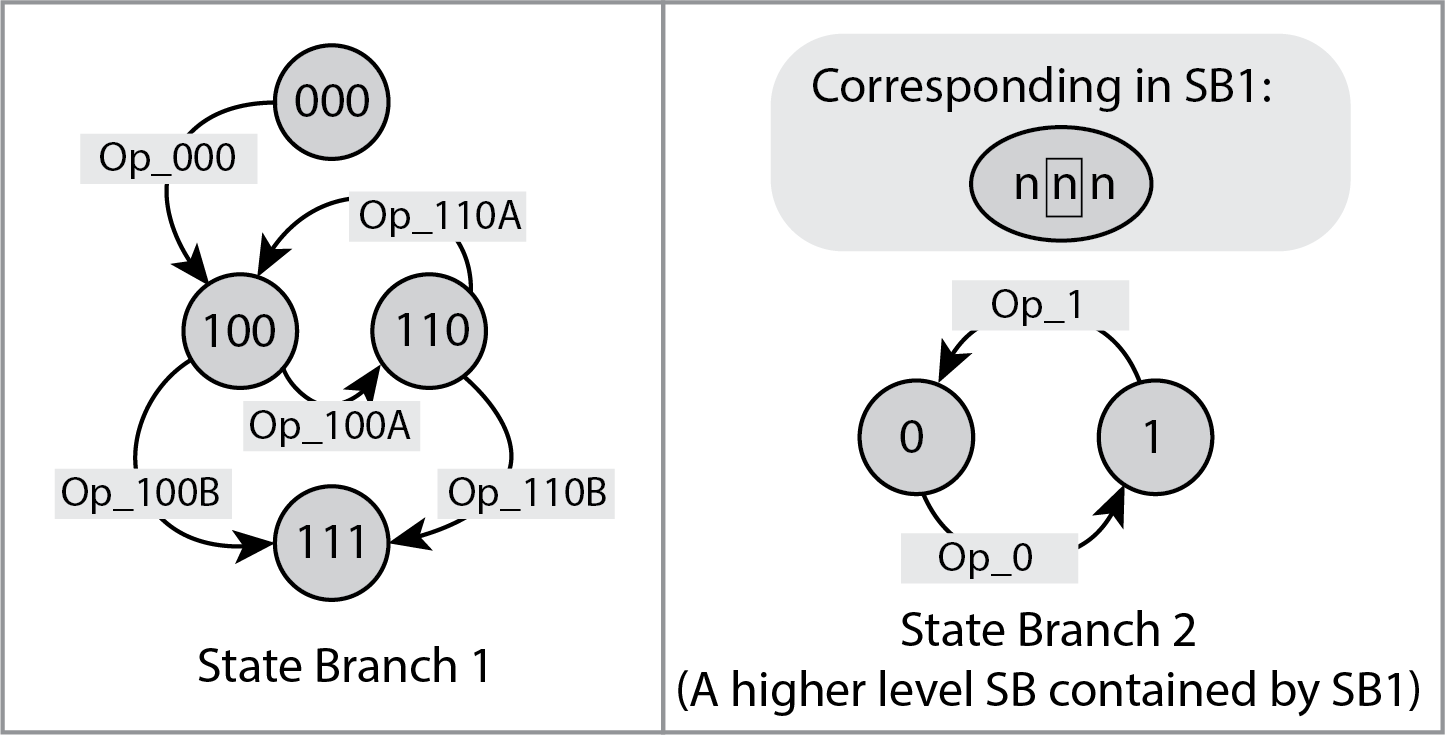}
    \caption{An example of hFSM. A higher level SB is a state digit in its lower level SB \cite{liu2023toward}.}
    \label{fig:hFSM}
\end{figure}

\begin{lstlisting}[language=json, linewidth=8.5cm]
{
    "_comments": {
        "HEADER": "",
        "INITIAL": "the initial state of the SB, default the first state",
        "ACTIVATING": "the activating state of the SB (required for higher level SB), default the last state",
        "NUM_FACTS": "the number of facts (state digits)",
        "SUB_SBS": "the higher level SB it contains and its corresponding state digit"
    },

    "_comment_note": {
        "UNDERSCORE_FIELDS": "The fields with the name starts with underscore is skipped"
    },

    "SB1": {
        "HEADER": {
            "INITIAL": "State000",
            "NUM_FACTS": 3,
            "SUB_SBS": {
                "SB2": 1
            }
        },

        "State000": {
            "Operation000": "State100"
        },

        "State100": {
            "Operation100A": "State110",
            "Operation100B": "State111"
        },

        "State110": {
            "Operation110A": "State100",
            "Operation110B": "State111"
        },

        "State111": { }
    },

    "SB2": {
        "HEADER": {
            "INITIAL": "State0",
            "ACTIVATING": "State1",
            "NUM_FACTS": 1
        },

        "State0": {
            "Operation0": "State1"
        },

        "State1": {
            "Operation1": "State0"
        }
    }
    
}
\end{lstlisting}

\end{appendices}

\end{document}